# On the Practical use of Variable Elimination in Constraint Optimization Problems: 'Still-life' as a Case Study

**Javier Larrosa**                                              LARROSA@LSI.UPC.EDU
**Enric Morancho**                                                ENRICM@AC.UPC.EDU
**David Niso**                                                 NISO57@CASAL.UPC.EDU
*Universitat Politecnica de Catalunya*
*Jordi Girona 1-3, 08034 Barcelona, Spain*

## Abstract

*Variable elimination* is a general technique for *constraint processing*. It is often discarded because of its high space complexity. However, it can be extremely useful when combined with other techniques. In this paper we study the applicability of variable elimination to the challenging problem of finding *still-lifes*. We illustrate several alternatives: variable elimination as a stand-alone algorithm, interleaved with search, and as a source of good quality lower bounds. We show that these techniques are the best known option both theoretically and empirically. In our experiments we have been able to solve the $n = 20$ instance, which is far beyond reach with alternative approaches.

## 1. Introduction

Many problems arising in domains such as *resource allocation* (Cabon, de Givry, Lobjois, Schiex, & Warners, 1999), *combinatorial auctions* (Sandholm, 1999), *bioinformatics* and *probabilistic reasoning* (Pearl, 1988) can be naturally modeled as *constraint satisfaction* and *optimization problems*. The two main solving schemas are *search* and *inference*. *Search* algorithms constitute the usual solving approach. They transform a problem into a set of subproblems by selecting one variable and instantiating it with its different alternatives. Subproblems are solved applying recursively the same transformation rule. The recursion defines a search tree that is normally traversed in a depth-first manner, which has the benefit of requiring only polynomial space. The practical efficiency of search algorithms greatly depends on their ability to detect and prune redundant subtrees. In the worst-case, search algorithms need to explore the whole search tree. Nevertheless, pruning techniques make them much more effective.

*Inference* algorithms (also known as *decomposition* methods) solve a problem by a sequence of transformations that reduce the problem size, while preserving its optimal cost. A well known example is *bucket elimination* (BE, also known as *variable elimination*) (Bertele & Brioschi, 1972; Dechter, 1999). The algorithm proceeds by selecting one variable at a time and replacing it by a new constraint which summarizes the effect of the chosen variable. The main drawback of BE is that new constraints may have large arities which require exponentially time and space to process and store. However, a nice property of BE is that its worst-case time and space complexities can be tightly bounded by a structural parameter called induced width. The exponential space complexity limits severely the algorithm's practical usefulness. Thus, in the constraint satisfaction community variable elimination is often disregarded.

In this paper we consider the challenging problem of finding *still-lifes* which are stable patterns of maximum density in the game of *life*. This academic problem has been recently included in the *CSPlib* repository[1] and a dedicated web page[2] has been set to maintain up-to-date results. In Bosch and Trick (2002), the still-life problem is solved using two different approaches: *integer programming* and *constraint programming*, both of them based on search. None of them could solve up to the $n = 8$ problem within reasonable time. Their best results were obtained with a hybrid approach which combines the two techniques and exploits the problem symmetries in order to reduce the search space. With their algorithm, they solved the $n = 15$ case in about 8 days of *cpu*. Smith (2002) proposed an interesting alternative using pure constraint programming techniques, and solving the problem in its dual form. In her work, Smith could not improve the $n = 15$ limit. Although not explicitly mentioned, these two works use algorithms with worst-case time complexity $O(2^{(n^2)})$.

In this paper we show the usefulness of variable elimination techniques. First we apply plain BE. Against what could be expected, we observe that BE is competitive with state-of-the-art alternatives. Next, we introduce a more sophisticated algorithm that combines search and variable elimination (following the ideas of Larrosa & Dechter, 2003) and uses a lower bound based on mini-buckets (following the ideas of Kask & Dechter, 2001). With our algorithm, we solve in one minute the $n = 15$ instance. We have been able to solve up to the $n = 20$ instance, which was far beyond reach with previous techniques. For readability reasons, we only describe the main ideas and omit algorithmic details.[3]

The structure of the paper is the following: In the next Section we give some preliminary definitions. In Section 3 we solve the problem with plain BE. In Section 4 we introduce the hybrid algorithm with which we obtained the results reported in Section 5. In Section 6 we discuss how the ideas explored in this article can be extended to other domains. Besides, we report additional experimental results. Finally, Section 7 gives some conclusions and lines of future work.

## 2. Preliminaries

In this Section we first define the still-life problem. Next, we define the weighted CSP framework and formulate the still-life as a weighted CSP. Finally, we review the main solving techniques for weighted CSPS.

### 2.1 Life and Still-Life

The game of *life* (Gardner, 1970) is played over an infinite checkerboard, where each square is called a *cell*. Each cell has eight *neighbors*: the eight cells that share one or two corners with it. The only player places checkers on some cells. If there is a checker on it, the cell is *alive*, else it is *dead*. The state of the board evolves iteratively according to the following three rules: (1) if a cell has exactly two living neighbors then its state remains the same

---

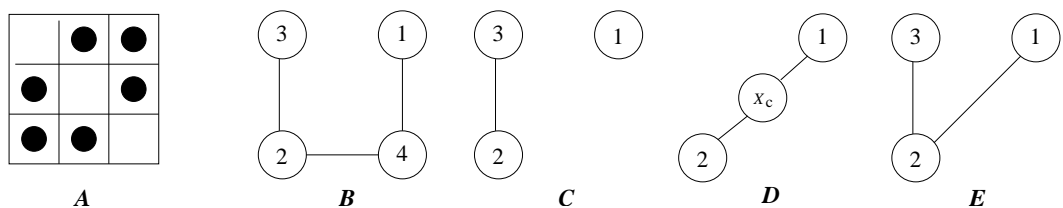

Figure 1: **A:** A $3 \times 3$ still-life. **B:** constraint graph of a simple WCSP instance with four variables and three cost functions. **C:** the constraint graph after *assigning* variable $x_4$. **D:** the constraint graph after *clustering* variables $x_3$ and $x_4$. **E:** the constraint graph after *eliminating* variable $x_4$.

in the next iteration, (2) if a cell has exactly three living neighbors then it is alive in the next iteration and (3) if a cell has fewer than two or more than three living neighbors, then it is dead in the next iteration. Although defined in terms of extremely simple rules, the game of life has proven mathematically rich and it has attracted the interest of both mathematicians and computer scientists.

The *still-life* problem $SL(n)$ consist on finding a $n \times n$ stable pattern of maximum density in the game of life. All cells outside the pattern are assumed to be dead. Considering the rules of the game, it is easy to see that each cell $(i, j)$ must satisfy the following three conditions: (1) if the cell is alive, it must have exactly two or three living neighbors, (2) if the cell is dead, it must not have three living neighbors, and (3) if the cell is at the grid boundary (i.e, $i = 1$ or $i = n$ or $j = 1$ or $j = n$), it cannot be part of a sequence of three consecutive living cells along the boundary. The last condition is needed because three consecutive living cells at a boundary would produce living cells outside the grid.

**Example 1** *Figure 1.A shows a solution to $SL(3)$. It is easy to verify that all its cells satisfy the previous conditions, hence it is stable. The pattern is optimal because it has 6 living cells and no $3 \times 3$ stable pattern with more that 6 living cells exists.*

## 2.2 Weighted CSP

A *weighted constraint satisfaction problem* (WCSP) (Bistarelli, Montanari, & Rossi, 1997) is defined by a tuple $(X, D, \mathcal{F})$, where $X = \{x_1, \ldots, x_n\}$ is a set of *variables* taking values from their finite *domains* $D_i \in D$. $\mathcal{F}$ is a set of *weighted constraints* (*i.e.*, cost functions). Each $f \in \mathcal{F}$ is defined over a subset of variables, $var(f)$, called its *scope*. The objective function is the sum of all functions in $\mathcal{F}$,

$$F = \sum_{f \in \mathcal{F}} f$$

and the goal is to find the instantiation of variables that *minimizes* the objective function.

**Example 2** *Consider a WCSP with four variables $X = \{x_i\}_{i=1}^{4}$ with domains $D_i = \{0, 1\}$ and three cost functions: $f_1(x_1, x_4) = x_1 + x_4$, $f_2(x_2, x_3) = x_2 x_3$ and $f_3(x_2, x_4) = x_2 + x_4$.*

*The objective function is $F(x_1, x_2, x_3, x_4) = x_1 + x_4 + x_2 x_3 + x_2 + x_4$. Clearly, the optimal cost is $0$, which is obtained with every variable taking value $0$.*

Constraints can be given explicitly by means of tables, or implicitly as mathematical expressions or computing procedures. Infeasible partial assignments are specified by constraints that assign cost $\infty$ to them. The assignment of value $a$ to variable $x_i$ is noted $x_i = a$. A partial assignment is a tuple $t = (x_{i_1} = v_1, x_{i_2} = v_2, \cdots, x_{i_j} = v_j)$. The extension of $t$ to $x_i = a$ is noted $t \cdot (x_i = a)$. WCSPs instances are graphically depicted by means of their *interaction* or *constraint graph*, which has one node per variable and one edge connecting any two nodes that appear in the same scope of some cost function. For instance, Figure 1.$B$ shows the constraint graph of the problem in the previous example.

## 2.3 Overview of Some Solving Techniques

In this Subsection we review some solving techniques widely used when reasoning with constraints.

### 2.3.1 SEARCH

WCSPs are typically solved with depth-first search. Search algorithms can be defined in terms of *instantiating* functions,

**Definition 1** *Let $P = (X, D, \mathcal{F})$ a WCSP instance, $f$ a function in $\mathcal{F}$, $x_i$ a variable in $var(f)$, and $v$ a value in $D_i$. Instantiating $f$ with $x_i = v$ is a new function with scope $var(f) - \{x_i\}$ which returns for each tuple $t$, $f(t \cdot (x_i = v))$. Instantiating $P$ with $x_i = v$ is a new problem $P \mid_{x_i = v} = (X - \{x_i\}, D - \{D_i\}, \mathcal{F}')$, where $\mathcal{F}'$ is obtained by instantiating all the functions in $\mathcal{F}$ that mention $x_i$ with $x_i = v$.*

For instance, instantiating the problem of Example 2 with $x_4 = 1$, produces a new problem with three variables $\{x_i\}_{i=1}^3$ and three cost functions: $f_1(x_1, x_4 = 1) = x_1 + 1$, $f_2(x_2, x_3) = x_2 x_3$ and $f_3(x_2, x_4 = 1) = x_2 + 1$. Figure 1.$C$ shows the corresponding constraint graph, obtained from the original graph by removing the instantiated variable $x_4$ and all adjacent edges. Observe that the new graph depends on the instantiated variable, but does not depend on the value assigned to it.

Search algorithms transform the current problem $P$ into a set of subproblems. Usually it is done by selecting one variable $x_i$ which is instantiated with its different domain values $(P \mid_{x_i = v_1}, P \mid_{x_i = v_2}, \cdots, P \mid_{x_i = v_d})$. This transformation is called *branching*. In each subproblem the same process is recursively applied, which defines a tree of subproblems. Search algorithms expand subproblems until a trivial case is achieved: there is no variable left, or a pruning condition is detected. In optimization problems, pruning conditions are usually defined in terms of lower and upper bounds. Search keeps the cost of the best solution so far, which is an *upper bound* of the optimal cost. At each node, a *lower bound* of the best cost obtainable underneath is computed. If the lower bound is greater than or equal to the upper bound, it is safe to backtrack.

The size of the search tree is $O(d^n)$ (being $d$ the size of the largest domain) which bounds the time complexity. If the tree is traversed depth-first, the space complexity is polynomial.

### 2.3.2 CLUSTERING

A well-known technique for constraint processing is *clustering* (Dechter & Pearl, 1989). It merges several variables into one *meta-variable*, while preserving the problem semantics. Clustering variables $x_i$ and $x_j$ produces meta-variable $x_k$, whose domain is $D_i \times D_j$. Cost functions must be accordingly clustered. For instance, in the problem of Example 2, clustering variables $x_3$ and $x_4$ produces variable $x_c$ with domain $D_c = \{(0,0), (0,1), (1,0), (1,1)\}$. Cost functions $f_2$ and $f_3$ are clustered into $f_c(x_2, x_c) = f_2 + f_3$. With the new variable notation $f_c = x_2 x_c[1] + x_2 + x_c[2]$, where $x_c[i]$ denotes the $i$-th component of $x_c$. Function $f_1$ needs to be reformulated as $f_1(x_1, x_c) = x_1 + x_c[2]$. The constraint graph of the resulting problem is obtained by merging the clustered variables and connecting the *meta-node* with all nodes that were adjacent to some of the clustered variables. Figure 1.$D$ shows the constraint graph after the clustering of $x_3$ and $x_4$. The typical use of clustering is to transform a cyclic constraint graph into an acyclic one, which can be solved efficiently thereafter.

### 2.3.3 VARIABLE ELIMINATION

*Variable elimination* is based on the following two operations,

**Definition 2** *The* sum *of two functions $f$ and $g$, noted $(f + g)$, is a new function with scope $var(f) \cup var(g)$ which returns for each tuple the sum of costs of $f$ and $g$,*

$$(f + g)(t) = f(t) + g(t)$$

**Definition 3** *The* elimination *of variable $x_i$ from $f$, noted $f \Downarrow x_i$, is a new function with scope $var(f) - \{x_i\}$ which returns for each tuple $t$ the cost of the best extension of $t$ to $x_i$,*

$$(f \Downarrow x_i)(t) = \min_{a \in D_i} \{f(t \cdot (x_i = a))\}$$

Observe that when $f$ is a unary function (*i.e.*, arity one), eliminating the only variable in its scope produces a constant.

**Definition 4** *Let $P = (X, D, \mathcal{F})$ be a WCSP instance. Let $x_i \in X$ be an arbitrary variable and let $B_i$ be the set of all cost functions having $x_i$ in their scope ($B_i$ is called the bucket of $x_i$). We define $g_i$ as*

$$g_i = (\sum_{f \in B_i} f) \Downarrow x_i$$

*The elimination of $x_i$ transforms $P$ into a new problem $P \Downarrow_{x_i} = \{X - \{x_i\}, D - \{D_i\}, (\mathcal{F} - B_i) \cup \{g_i\}\}$. In words, $P \Downarrow_{x_i}$ is obtained by replacing $x_i$ and all the functions in its bucket by $g_i$.*

$P$ and $P \Downarrow_{x_i}$ have the same optimal cost because, by construction, $g_i$ *compensates* the absence of $x_i$. The constraint graph of $P \Downarrow_{x_i}$ is obtained by forming a clique with all the nodes adjacent to node $x_i$ and then removing $x_i$ and all its adjacent edges. For example, eliminating $x_4$ in the problem of Example 2 produces a new problem with three variables $\{x_i\}_{i=1}^3$ and two cost functions: $f_2$ and $g_4$. The scope of $g_4$ is $\{x_1, x_2\}$ and it is defined as, $g_4 = (f_1 + f_3) \Downarrow x_4 = (x_1 + x_4 + x_2 + x_4) \Downarrow x_4 = x_1 + x_2$. Figure 1.$D$ shows the constraint graph after the elimination.

In the previous example, the new function $g_4$ could be expressed as a mathematical expression. Unfortunately, in general, the result of summing functions or eliminating variables cannot be expressed intensionally, and new cost functions must be stored extensionally in tables. Consequently, the space complexity of computing $P \Downarrow_{x_i}$ is proportional to the number of entries of $g_i$, which is: $\Theta(\prod_{x_j \in var(g_i)} |D_j|)$. Since $x_j \in var(g_i)$ iff $x_j$ is adjacent to $x_i$ in the constraint graph, the previous expression can be rewritten as $\Theta(\prod_{x_j \in N(i,G_P)} |D_j|)$, where $G_P$ is the constraint graph of $P$ and $N(i, G_P)$ is the set of neighbors of $x_i$ in $G_P$. The time complexity of computing $P \Downarrow_{x_i}$ is its space complexity multiplied by the cost of computing each entry of $g_i$.

*Bucket elimination* (BE) works in two phases. In the first phase, it eliminates variables one at a time in reverse order. In the elimination of $x_i$, the new $g_i$ function is computed and added to the corresponding bucket. The elimination of $x_1$ produces an empty-scope function (*i.e.*, a constant) which is the optimal cost of the problem. In the second phase, BE considers variables in increasing order and generates the optimal assignment of variables. The time and space complexity of BE is exponential on a structural parameter from the constraint graph, called induced width, which captures the maximum arity among all the $g_i$ functions. Without any additional overhead BE can also compute the number of optimal solutions (see Dechter, 1999, for details).

### 2.3.4 SUPER-BUCKETS

In some cases, it may be convenient to eliminate a set of variables simultaneously (Dechter & Fatah, 2001). The elimination of the set of variables $Y$ is performed by collecting in $B_Y$ the set of functions mentioning at least one variable of $Y$. Variables in $Y$ and functions in $B_Y$ are replaced by a new function $g_Y$ defined as,

$$g_Y = ( \sum_{f \in B_Y} f) \Downarrow Y$$

The set $B_Y$ is called a *super-bucket*. Note that the elimination of $Y$ can be seen as the clustering of its variables into a meta-variable $x_Y$ followed by its elimination.

### 2.3.5 MINI-BUCKETS

When the space complexity of BE is too high, an approximation, called *mini buckets* (Dechter & Rish, 2003), can be used. Consider the elimination of $x_i$, with its associated bucket $B_i = \{f_{i_1}, \ldots, f_{i_k}\}$. BE would compute,

$$g_i = ( \sum_{f \in B_i} f) \Downarrow x_i$$

The time and space complexity of this computation depends on the arity of $g_i$. If it is beyond our available resources, we can partition bucket $B_i$ into so-called *mini-buckets* $B_{i_1}, \ldots, B_{i_k}$ where the number of variables in the scopes of each mini-bucket is bounded by a parameter. Then we can compute,

$$g_{i_j} = ( \sum_{f \in B_{i_j}} f) \Downarrow x_i, \quad j = 1..k$$

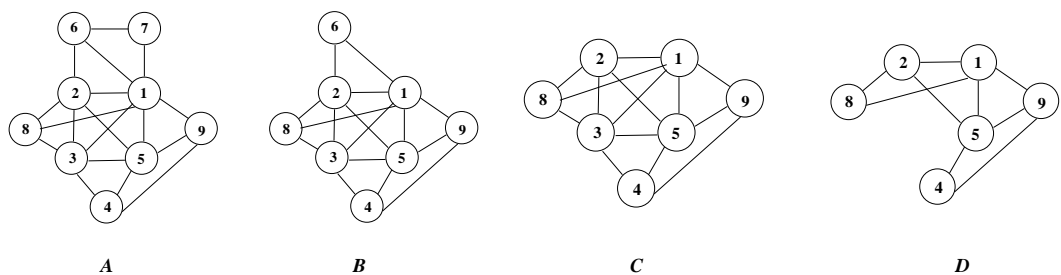

Figure 2: A constraint graph and its evolution over a sequence of variable eliminations and instantiations.

where each $g_{i_j}$ has a bounded arity. Since,

$$\underbrace{(\sum_{f \in B_i} f) \Downarrow x_i}_{g_i} \geq \sum_{j=1}^{k} \underbrace{(\sum_{f \in B_{i_j}} f) \Downarrow x_i}_{g_{i_j}}$$

the elimination of variables using mini-buckets yields a lower bound of the actual optimal cost.

### 2.3.6 COMBINING SEARCH AND VARIABLE ELIMINATION

When plain BE is too costly in space, we can combine it with search (Larrosa & Dechter, 2003). Consider a WCSP whose constraint graph is depicted in Figure 2.A. Suppose that we want to eliminate a variable but we do not want to compute and store constraints with arity higher than two. Then we can only take into consideration variables connected to at most two variables. In the example, variable $x_7$ is the only one that can be selected. Its elimination transforms the problem into another one whose constraint graph is depicted in Figure 2.B. Now $x_6$ has its degree decreased to two, so it can also be eliminated. The new constraint graph is depicted in Figure 2.C. At this point, every variable has degree greater than two, so we switch to a search schema which selects a variable, say $x_3$, branches over its values and produces a set of subproblems, one for each value in its domain. All of them have the same constraint graph, depicted in Figure 2.D. For each subproblem, it is possible to eliminate variable $x_8$ and $x_4$. After their elimination it is possible to eliminate $x_2$ and $x_9$, and subsequently $x_5$ and $x_1$. Eliminations after branching have to be done at every subproblem since the new constraints with which the eliminated variables are replaced differ from one subproblem to another. In the example, only one branching has been made. Therefore, the elimination of variables has reduced the search tree size from $d^9$ to $d$, where $d$ is the size of the domains. In the example, we bounded the arity of the new constraints to two, but it can be generalized to an arbitrary value.

## 3. Solving Still-life with Variable Elimination

$SL(n)$ can be easily formulated as a WCSP. The most natural formulation associates one variable $x_{ij}$ with each cell $(i, j)$. Each variable has two domain values. If $x_{ij} = 0$ the cell is dead, if $x_{ij} = 1$ it is alive. There is a cost function $f_{ij}$ for each variable $x_{ij}$. The scope of $f_{ij}$ is $x_{ij}$ and all its neighbors. It evaluates the stability of $x_{ij}$: if $x_{ij}$ is unstable given its neighbors, $f_{ij}$ returns $\infty$; else $f_{ij}$ returns $1 - x_{ij}$.[4] The objective function to be minimized is,

$$F = \sum_{i=1}^{n} \sum_{j=1}^{n} f_{ij}$$

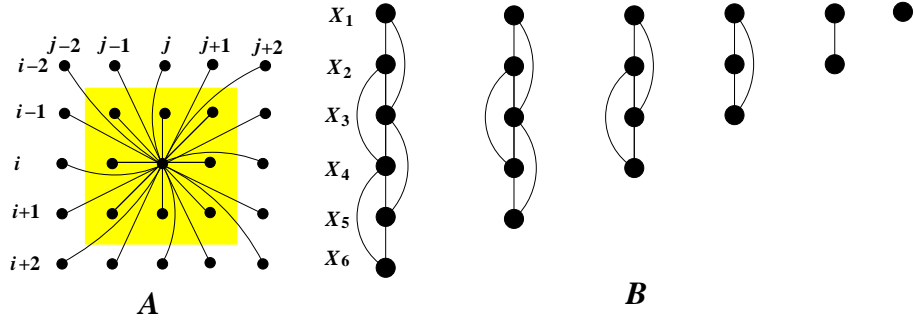

Figure 3: *A*: Structure of the constraint graph of $SL(n)$. The node in the center, associated to cell $(i, j)$, is linked to all cells it interacts with. The shadowed area indicates the scope of $f_{ij}$. *B (left)*: Constraint graph of $SL(6)$ after clustering cells into row variables. *B (from left to right*: Evolution of the constraint graph during the execution of BE.

If the instantiation $X$ represents an unstable pattern, $F(X)$ returns $\infty$; else it returns the number of dead cells. $f_{ij}$ can be stored as a table with $2^9$ entries and evaluated in constant time.

Figure 3.*A* illustrates the structure of the constraint graph of $SL(n)$. The picture shows an arbitrary node $x_{ij}$ linked to all the nodes it interacts with. For instance, there is an edge between $x_{ij}$ and $x_{i,j+1}$ because $x_{i,j+1}$ is a neighbor of $x_{ij}$ in the grid and, consequently, both variables are in the scope of $f_{ij}$. There is an edge between $x_{ij}$ and $x_{i-1,j-2}$ because both cells are neighbors of $x_{i-1,j-1}$ in the grid and, therefore, both appear in the scope of $f_{i-1,j-1}$. The shadowed area represents the scope of $f_{ij}$ (namely, $x_{ij}$ and all its neighbors). The complete graph is obtained by extending this connectivity pattern to all nodes in the graph.

For the sake of clarity, we use an equivalent but more compact $SL(n)$ formulation that makes BE easier to describe and implement: we *cluster* all variables of each row into a single meta-variable. Thus, $x_i$ denotes the state of cells in the $i$-th row (namely, $x_i = (x_{i1}, x_{i2}, \ldots, x_{in})$ with $x_{ij} \in \{0, 1\}$). Accordingly, it takes values over the sequences of $n$ bits or, equivalently, over the natural numbers in the interval $[0..2^n - 1]$. Cost functions are accordingly clustered: there is a cost function $f_i$ associated with each row $i$, defined as,

$$f_i = \sum_{j=1}^{n} f_{ij}$$

---

4. Recall that, as a WCSP, the task is to *minimize* the number of dead cells. Therefore, we give cost 1 to dead cells and cost 0 to living cells.

For internal rows, the scope of $f_i$ is $\{x_{i-1}, x_i, x_{i+1}\}$. The cost function of the top row, $f_1$, has scope $\{x_1, x_2\}$. The cost function of the bottom row, $f_n$, has scope $\{x_{n-1}, x_n\}$. If there is some unstable cell in $x_i$, $f_i(x_{i-1}, x_i, x_{i+1}) = \infty$. Else, it returns the number of dead cells in $x_i$. Evaluating $f_i$ is $\Theta(n)$ because all the bits of the arguments need to be checked. The new, equivalent, objective function is,

$$F = \sum_{i=1}^{n} f_i$$

Figure 3.$B$ (left) shows the constraint graph of $SL(6)$ with this formulation. An arbitrary variable $x_i$ is connected with the two variables above and the two variables below. The sequential structure of the constraint graph makes BE very intuitive. It eliminates variables in decreasing orders. The elimination of $x_i$ produces a new function $g_i = (f_{i-1} + g_{i+1}) \downarrow x_i$ with scope $\{x_{i-2}, x_{i-1}\}$. Figure 3.$B$ (from left to right) shows the evolution of the constraint graph along the elimination of its variables. Formally, BE applies a recursion that transforms subproblem $P$ into $P \Downarrow_{x_i}$, where $x_i$ is the variable in $P$ with the highest index. It satisfies the following property,

**Property 1** *Let $g_i$ be the function added by BE to replace $x_i$. Then $g_i(a, b)$ is the cost of the best extension of $(x_{i-2} = a,\ x_{i-1} = b)$ to the eliminated variables $(x_i, \ldots, x_n)$. Formally,*

$$
\begin{aligned}
g_i(a, b) \quad = \quad & \min_{v_i \in D_i, \ldots, v_n \in D_n} \{ f_{i-1}(a, b, v_i) + f_i(b, v_i, v_{i+1}) + \\
& + f_{i+1}(v_i, v_{i+1}, v_{i+2}) + \ldots \\
& + f_{n-1}(v_{n-2}, v_{n-1}, v_n) + f_n(v_{n-1}, v_n) \}
\end{aligned}
$$

If $g_i(a, b) = \infty$, it means that the pattern $a, b$ cannot be extended to the inferior rows with a stable pattern. If $g_i(a, b) = k$ (with $k \neq \infty$), it means that $a, b$ can be extended and the optimal extension has $k$ dead cells from $x_{i-1}$ to $x_n$.

The space complexity of BE $\Theta(n \times 2^{2n})$, due to the space required to store $n$ functions $g_i$ extensionally ($2^n \times 2^n$ entries each). Regarding time, computing each entry of $g_i$ has cost $\Theta(n \times 2^n)$ (finding the minimum of $2^n$ alternatives, the computation of each one is $\Theta(n)$). Since each $g_i$ has $2^{2n}$ entries, the total time complexity is $\Theta(n^2 \times 2^{3n})$. Observe that solving $SL(n)$ with BE is an exponential improvement over search algorithms, which have time complexity $O(2^{n^2})$.

Table 4 reports some empirical results. They were obtained with a 2 Ghz Pentium IV machine with 2 Gb of memory. The first columns reports the problem size, the second reports the optimal cost as the number of dead cells (in parenthesis, the number of living cells), the third column reports the number of optimal solutions. We count as different two solutions even if one can be transformed to the other through a problem symmetry. The fourth column reports the CPU time of BE in seconds. The fifth, sixth and seventh columns report the results obtained with the three approaches tried by Bosch and Trick (2002):[5] constraint programming (CP), integer programming (IP), and a more sophisticated algorithm (CP/IP) which combines CP and IP, and exploits the problem symmetries.

---

5. The corresponding OPL code is available at `http://mat.gsia.cmu.edu/LIFE`.

| $n$ | opt | n. sol. | BE | CP | IP | CP/IP |
|-----|-----|---------|-----|-----|-----|-------|
| 5 | 9(16) | 1 | 0 | 0 | 0 | 0 |
| 6 | 18(18) | 48 | 0 | 0 | 1 | 0 |
| 7 | 21(28) | 2 | 0 | 4 | 3 | 0 |
| 8 | 28(36) | 1 | 0 | 76 | 26 | 2 |
| 9 | 38(43) | 76 | 4 | > 600 | > 600 | 20 |
| 10 | 46(54) | 3590 | 27 | * | * | 60 |
| 11 | 57(64) | 73 | 210 | * | * | 153 |
| 12 | 68(76) | 129126 | 1638 | * | * | 11536 |
| 13 | 79(90) | 1682 | 13788 | * | * | 12050 |
| 14 | 92(104) | 11 | $10^5$ | * | * | $5 \times 10^5$ |
| 15 | 106(119) | ? | * | * | * | $7 \times 10^5$ |

Figure 4: Experimental results of four different algorithms on the still-life problem. Times are in seconds.

It can be observed that BE clearly outperforms CP and IP by orders of magnitude. The $n = 14$ case is the largest instance that we could solve due to exhausting the available space. Comparing BE with CP/IP, we observe that there is no clear winner. An additional observation is that BE scales up very regularly, each execution requiring roughly eight times more time and four times more space than the previous, which is in clear accordance with the algorithm complexity.

## 4. Combining Search and Variable Elimination

One way to overcome the high space complexity of BE is to combine search and variable elimination in a hybrid approach HYB (Larrosa & Schiex, 2003). The idea is to use search (i.e, instantiations) in order to *break* the problem into independent smaller parts where variable elimination can be efficiently performed.

Let us reformulate the problem in a more convenient way for the hybrid algorithm. For the sake of simplicity and without loss of generality consider that $n$ is even. We cluster row variables into three meta-variables: $x_i^C$ denotes the two central cells of row $i$, $x_i^R$ and $x_i^L$ denote the $\frac{n}{2} - 1$ remaining cells on the right and left, respectively (see Figure 5.A). Consequently, $x_i^C$ takes values in the range $[0..3]$, $x_i^L$ and $x_i^R$ take values in the range $[0..2^{\frac{n}{2}-1} - 1]$. Cost functions are accordingly clustered,

$$f_i^L = \sum_{j=1}^{\frac{n}{2}} f_{ij}, \quad f_i^R = \sum_{j=\frac{n}{2}+1}^{n} f_{ij}$$

The new, equivalent, objective function is,

$$F = \sum_{i=1}^{n} (f_i^L + f_i^R)$$

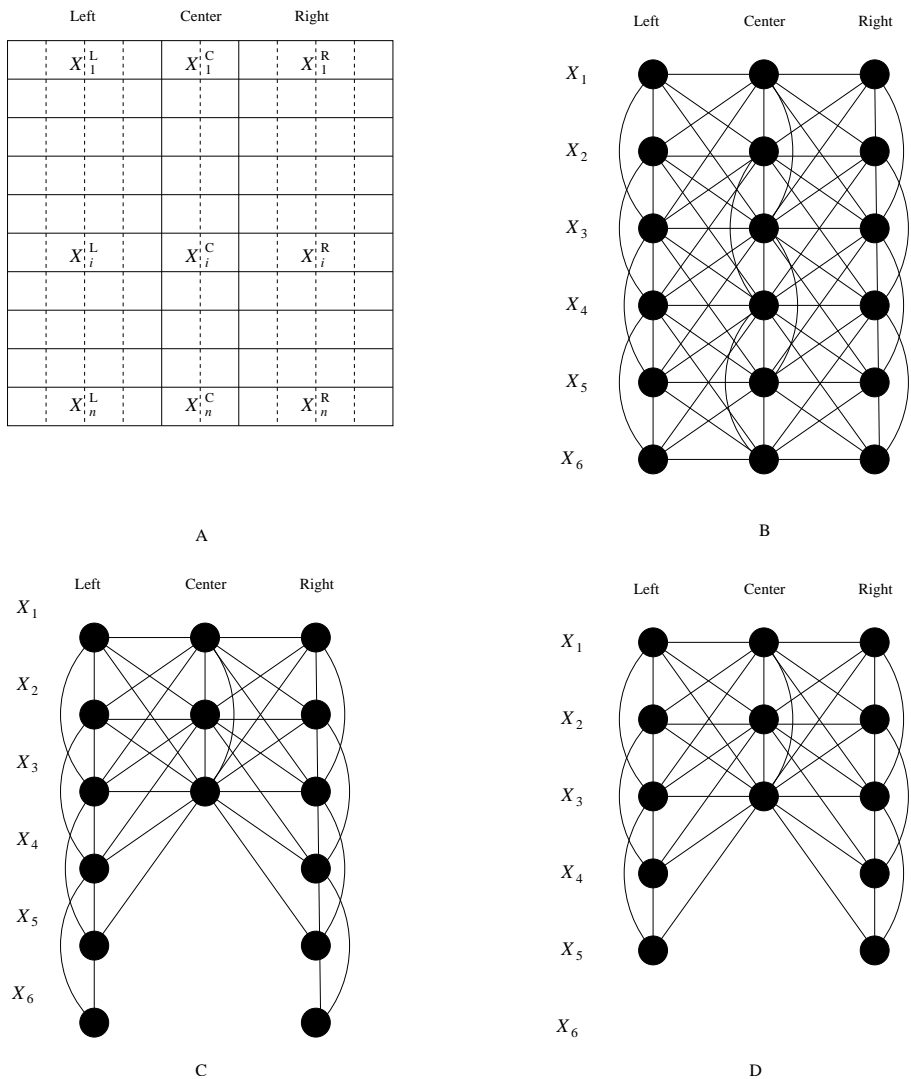

Figure 5: Formulation of $SL(n)$ used by the hybrid algorithm. *A*: Each row is clustered into three variables. *B*: Constraint graph of $SL(6)$. *C*: Constraint graph after the assignment of $x_n^C$, $x_{n-1}^C$ and $x_{n-2}^C$. *D*: Constraint graph after the elimination of $x_n^L$ and $x_n^R$.

The scopes of internal row functions, $f_i^L$ and $f_i^R$, are $\{x_{i-1}^L, x_{i-1}^C, x_i^L, x_i^C, x_{i+1}^L, x_{i+1}^C\}$ and $\{x_{i-1}^C, x_{i-1}^R, x_i^C, x_i^R, x_{i+1}^C, x_{i+1}^R\}$. Top functions $f_1^L$ and $f_1^R$ have scopes $\{x_1^L, x_1^C, x_2^L, x_2^C\}$ and $\{x_1^C, x_1^R, x_2^C, x_2^R\}$. Bottom functions $f_n^L$ and $f_n^R$ have scopes $\{x_n^L, x_{n-1}^C, x_n^L, x_n^C\}$ and $\{x_{n-1}^C, x_{n-1}^R, x_n^C, x_n^R\}$. Figure 5.$B$ shows the corresponding constraint graph. The importance of this formulation is that $x_i^L$ and $x_i^R$ are independent (i.e, there is no edge in the constraint graph connecting left and right variables).

The hybrid algorithm HYB searches over the central variables and eliminates the lateral variables. Variables are considered in decreasing order of their index. Thus, the algorithm starts instantiating $x_n^C$, $x_{n-1}^C$ and $x_{n-2}^C$, which produces a subproblem with the constraint graph shown in Figure 5.$C$. Observe that variable $x_n^L$ (respectively, $x_n^R$) is only connected with variables $x_{n-1}^L$ and $x_{n-2}^L$ (respectively, $x_{n-1}^R$ and $x_{n-2}^R$). Then it is eliminated producing a new function $g_n^L$ with scope $\{x_{n-2}^L, x_{n-1}^L\}$ (respectively, $g_n^R$ with scope $\{x_{n-2}^R, x_{n-1}^R\}$). Figure 5.$D$ shows the resulting constraint graph. Lateral variables have domains of size $2^{\frac{n}{2}-1}$. Hence, their elimination is space $\Theta(2^n)$ and time $\Theta(2^{3\frac{n}{2}})$. It is important to note that these eliminations are *subject* to the current assignment of $x_n^C$, $x_{n-1}^C$ and $x_{n-2}^C$. Therefore, they have to be recomputed when their value change. After the elimination of $x_n^L$ and $x_n^R$, the algorithm would assign variable $x_{n-3}^C$ which will make possible the elimination of $x_{n-1}^L$ and $x_{n-1}^R$, and so on. At an arbitrary level of search, the algorithm assigns $x_i^C$, which makes $x_{i+2}^L$ and $x_{i+2}^R$ independent of the central columns and only related to their two variables above. Then, it eliminates them by replacing the variables by functions $g_{i+2}^L$ and $g_{i+2}^R$ with scopes $\{x_i^L, x_{i+1}^L\}$ and $\{x_i^R, x_{i+1}^R\}$, respectively. Formally, HYB applies a recursion that transforms subproblem $P$ into 4 simpler subproblems $\{((P\mid_{x_i^C=v}) \Downarrow_{x_{i+2}^L}) \Downarrow_{x_{i+2}^R}\}_{v=0}^3$. It satisfies the following property,

**Property 2** *Let $g_i^L$ be the function computed by HYB used to replace variable $x_i^L$. Then $g_i^L(a,b)$ is the cost of the best extension of $(x_{i-2}^L = a,\ x_{i-1}^L = b)$ to eliminated variables $(x_i^L, \ldots, x_n^L)$, conditioned to the current assignment. Similarly, for the right side, $g_i^R(a,b)$ is the cost of the best extension of $(x_{i-2}^R = a,\ x_{i-1}^R = b)$ to eliminated variables $(x_i^R, \ldots, x_n^R)$, conditioned to the current assignment.*

A consequence of the previous Property is that the minimum $g_{i+2}^L(a,b)$ among all combinations of $a$ and $b$ is a lower bound of the best cost that can be obtained in the left part of the grid if we continue the current line of search. Therefore, $\min_{a,b}\{g_{i+2}^L(a,b)\} + \min_{a,b}\{g_{i+2}^R(a,b)\}$ is a valid lower bound of the current node and can be used for pruning purposes.

The space complexity of the algorithm is $\Theta(n \times 2^n)$, due to the $g_i^L$ and $g_i^R$ functions which need to be explicitly stored. The time complexity is $O(n \times 2^{3.5n})$, because $O(4^n)$ nodes may be visited ($n$ variables with domains of size 4) and the cost of processing each node is $\Theta(n \times 2^{3\frac{n}{2}})$ due to the variable eliminations.

Thus, comparing with BE, the time complexity increases from $\Theta(n^2 \times 2^{3n})$ to $O(n \times 2^{3.5n})$. This is the prize HYB pays for the space decrement from $\Theta(n \times 2^{2n})$ to $\Theta(n \times 2^n)$.

## 4.1 Refining the Lower Bound

It is well-known that the average-case efficiency of search algorithms depends greatly on the lower bound that they use. Our algorithm is using a poor lower bound based on the $g_i^L$ and $g_i^R$ functions, only.

Kask and Dechter (2001) proposed a general method to incorporate information from yet-unprocessed variables into the lower bound. Roughly, the idea is to run mini buckets (MB) prior search and save intermediate functions for future use. MB is executed using the reverse order in which search will instantiate the variables. When the execution of MB is completed, the search algorithm is executed. At each node, it uses mini-bucket functions as compiled look-ahead information. In this Subsection, we show how we have adapted this idea to $SL(n)$ and how we have integrated it into HYB.

Consider $SL(n)$ formulated in terms of *left*, *central* and *right* variables $(x_i^L, x_i^C, x_i^R)$. The *exact* elimination of the first row variables $(x_1^L, x_1^C, x_1^R)$ can be done using *super-bucket* $B_1 = \{f_1^L, f_1^R, f_2^L, f_2^R\}$ and computing the function,

$$h_1 = (f_1^L + f_1^R + f_2^L + f_2^R) \Downarrow \{x_1^L, x_1^C, x_1^R\}$$

The scope of $h_1$ is $\{x_2^L, x_2^C, x_2^R, x_3^L, x_3^C, x_3^R\}$. Using the mini-buckets idea, we partition the bucket into $B_1^L = \{f_1^L, f_2^L\}$ and $B_1^R = \{f_1^R, f_2^R\}$. Then, we approximate $h_1$ by two smaller functions $h_1^L$ and $h_1^R$,

$$h_1^L = (f_1^L + f_2^L) \Downarrow \{x_1^L, x_1^C\}$$
$$h_1^R = (f_1^R + f_2^R) \Downarrow \{x_1^C, x_1^R\}$$

The scopes of $h_1^L$ and $h_1^R$ are $\{x_2^L, x_2^C, x_3^L, x_3^C\}$ and $\{x_2^C, x_2^R, x_3^C, x_3^R\}$, respectively. The same idea is repeated row by row in increasing order. In general, processing row $i$, yields two functions,

$$h_i^L = (h_{i-1}^L + f_{i+1}^L) \Downarrow \{x_i^L, x_i^C\}$$
$$h_i^R = (h_{i-1}^R + f_{i+1}^R) \Downarrow \{x_i^C, x_i^R\}$$

The scopes of $h_i^L$ and $h_i^R$ are $\{x_{i+1}^L, x_{i+1}^C, x_{i+2}^L, x_{i+2}^C\}$ and $\{x_{i+1}^C, x_{i+1}^R, x_{i+2}^C, x_{i+2}^R\}$, respectively. By construction, $h_i^L(a, a', b, b')$ contains the cost of the best extension of $a, a', b, b'$ to processed variables $x_i^L, x_i^C, \ldots, x_1^L, x_1^C$ considering left functions only. We have the same property for $h_i^R(a', a, b', b)$ and right functions.

The complexity of MB is space $\Theta(n \times 2^n)$ and time $\Theta(n^2 \times 2^{1.5n})$. Since these complexities are smaller than the complexity of HYB, running this pre-process does not affect its overall complexity.

After MB is executed, HYB can use the information recorded in the $h_i^L$ and $h_i^R$ functions. Consider an arbitrary node in which HYB assigns $x_i^C$ and eliminates $x_{i+2}^L$ and $x_{i+2}^R$. Let $a$ and $b$ be domain values of variables $x_i^L$ and $x_{i+1}^L$. From Property 2 we have that $g_{i+2}^L(a, b)$ contains the best extension of $a, b$ that can be attained in the left part of rows $i + 1$ to $n$ as long as the current assignment $X^C$ is maintained. Additionally, we have that $h_{i-1}^L(a, x_i^C, b, x_{i+1}^C)$ contains the best extension of $a, b$ that can be attained in the left part of rows $i$ to 1. Therefore, $g_{i+2}^L(a, b) + h_{i-1}^L(a, x_i^C, b, x_{i+1}^C)$ is a lower bound for $a, b$ and $X^C$ of the left part of the grid. Consequently,

$$\mathtt{min}_{a,b \in [0..2^{\frac{n}{2}-1}-1]}\{g_{i+2}^L(a, b) + h_{i-1}^L(a, x_i^C, b, x_{i+1}^C)\}$$

is a lower bound of the left part of the grid for the current assignment. With the same reasoning on the right part we have that,

$$\min_{a,b\in[0..2^{\frac{n}{2}-1}-1]}\{g^L_{i+2}(a,b)+h^L_{i-1}(a,x^C_i,b,x^C_{i+1})\}+$$
$$+\min_{a,b\in[0..2^{\frac{n}{2}-1}-1]}\{g^R_{i+2}(a,b)+h^R_{i-1}(x^C_i,a,x^C_{i+1},b)\}$$

is a lower bound of the current assignment.

## 4.2 Refining the Upper Bound

The efficiency of the algorithm also depends on the initial value of the upper bound. A good upper bound facilitates pruning earlier in the search tree. Bosch and Trick (2002) suggested to modify $SL(n)$ by adding the additional constraint of considering symmetric patterns, only. Since the space of solutions becomes considerably smaller, the problem is presumably simpler. Clearly, the cost of an optimal symmetric stable pattern is an upper bound of the optimal cost of $SL(n)$. It has been observed that such upper bounds are very tight.

Since the motivation of our work is to use variable elimination techniques, we have considered still-lifes which are symmetric over a vertical reflection, because they can be efficiently solved using BE. The symmetric still-life problem $SSL(n)$ consists on finding a $n \times n$ stable pattern of maximum density in the game of life subject to a vertical reflection symmetry (namely, the state of cells $(i,j)$ and $(i,n-j+1)$ must be the same.[6]

Adapting BE to solve $SSL(n)$ is extremely simple: we only need to remove symmetrical values from the domains. Let us assume that $n$ is an even number (the odd case is similar). We represent a symmetric sequences of bits of length $n$ by considering the left side of the sequence (i.e, the first $n/2$ bits). The right part is implicit in the left part. Thus, we represent symmetrical sequences of $n$ bits as integers in the interval $[0..2^{\frac{n}{2}}-1]$. Reversing a sequence of bits $a$ is noted $\bar{a}$. Hence, if $a$ is a sequence of $n/2$ bits, $a \cdot \bar{a}$ is the corresponding symmetrical sequence of $n$ bits.

The complexity of BE, when applied to $SSL(n)$ is time $\Theta(n^2 \times 2^{1.5n})$ and space $\Theta(n \times 2^n)$. Therefore, executing it prior HYB and setting the upper bound with its optimal cost does not affect the overall complexity of the hybrid.

## 4.3 Further Exploitation of Symmetries

$SL(n)$ is a highly symmetric problem. For any stable pattern, it is possible to create an equivalent pattern by: ($i$) rotating the board by 90, 180 or 270 degrees, ($ii$) reflecting the board horizontally, vertically or along one diagonal or ($iii$) doing any combination of rotations and reflections.

Symmetries can be exploited at very different algorithmic levels. In general, we can save any computation whose outcome is equivalent to a previous computation due to a symmetry if we have kept its outcome. For instance, in MB it is not necessary to compute $h^R_i(a',a,b',b)$ because it is equal to $h^L_i(\overline{a'},\overline{a},\overline{b},\overline{b'})$ due to the vertical reflection symmetry. Another example occurs in HYB. Let $x^C_n = v_n, x^C_{n-1} = v_{n-1}, \ldots, x^C_i = v_i$ be the current assignment. The reversed assignment $x_n^C = \overline{v_n}, x_{n-1}^C = \overline{v_{n-1}}, \ldots, x_i^C = \overline{v_i}$ is equivalent due to the vertical reflection symmetry. Thus, if it has already been considered, the algorithm can backtrack. Our implementation uses these tricks and some others which we do not report because it would require a much lower level description of the algorithms.

---

6. Unlike Smith's (2002) work we cannot easily exploit a larger variety of symmetries such as rotations and diagonal reflections.

| $n$ | opt | opt-SSL | CP/IP | BE | HYB | HYB no LB | HYB no UB |
|---|---|---|---|---|---|---|---|
| 13 | 79(90) | 79 | 12050 | 13788 | 2 | 2750 | 2 |
| 14 | 92(104) | 92 | $5 \times 10^5$ | $10^5$ | 2 | 7400 | 3 |
| 15 | 106(119) | 106 | $7 \times 10^5$ | * | 58 | $2 \times 10^5$ | 61 |
| 16 | 120(136) | 120 | * | * | 7 | $6 \times 10^5$ | 49 |
| 17 | 137(152) | 137 | * | * | 1091 | * | 2612 |
| 18 | 153(171) | 154 | * | * | 2029 | * | 2311 |
| 19 | 171(190) | 172 | * | * | 56027 | * | 56865 |
| 20 | 190(210) | 192 | * | * | $2 \times 10^5$ | * | $2 \times 10^5$ |
| 22 | ? | 232 | * | * | * | * | * |
| 24 | ? | 276 | * | * | * | * | * |
| 26 | ? | 326 | * | * | * | * | * |
| 28 | ? | 378 | * | * | * | * | * |

Figure 6: Experimental results of three different algorithms on the still-life problem. Times are in seconds.

## 5. Experimental Results

Figure 6 shows the empirical performance of our hybrid algorithm. The first column contains the problem size. The second column contains the optimal value as the number of dead cells (in parenthesis the corresponding number of living cells). The third column contains the optimal value of the symmetrical problem $SSL(n)$, obtained by executing BE. It can be observed that $SSL(n)$ provides very tight upper bounds to $SL(n)$. The fourth column reports the time obtained with the CP/IP algorithm (Bosch & Trick, 2002). The fifth column reports times obtained with BE. The sixth column contains times obtained with our hybrid algorithm HYB. As it can be seen, the performance of HYB is spectacular. The $n = 14$ and $n = 15$ instances, which require several days of CPU, are solved by HYB in a few seconds. Instances up to $n = 18$ are solved in less than one hour. The largest instance that we can solve is $n = 20$, which requires about two days of CPU (Figure 7 shows the optimal $n = 19$ and $n = 20$ still-lifes). Regarding space, our computer can handle executions of HYB up to $n = 22$. However, neither the $n = 21$ nor the $n = 22$ instance could be solved within a week of CPU. It may seem that solving the $n = 20$ instance is a petty progress with respect previous results on the problem. This is clearly not the case. The search space of the $n = 15$ and $n = 20$ instances have size $2^{15^2} = 2^{225}$ and $2^{20^2} = 2^{400}$, respectively. Thus, we have been able to solve a problem with a search space $2^{175}$ times larger than before. Since BE scales up very regularly, we can accurately predict that it would require 4000 Gb of memory and about 7 centuries to solve the $n = 20$ instance.

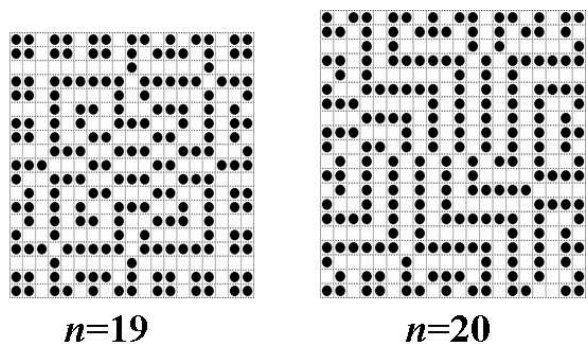

Figure 7: Maximum density still-lifes for $n = 19$ and $n = 20$.

Since HYB combines several techniques, it is interesting to assess the impact of each one. The seventh column reports times obtained with HYB without using mini-buckets information in the lower bound. As can be seen, the algorithm is still better than plain BE, but it performance is dramatically affected. The information gathered during the preprocess improves the quality of the lower bound and anticipates pruning. Finally, the eighth column reports times obtained with HYB without having the upper bound initialized to $SSL(n)$. In this case we see that the importance of this technique is quite limited. The reason is that HYB, even with a bad initial upper bound, finds the optimum very rapidly and, after that moment, the quality of the initial upper bound becomes irrelevant.

## 6. Extension to Other Domains

The $SL(n)$ problem has a very well defined structure, and the hybrid algorithm that we have proposed makes an *ad hoc* exploitation of it. It is easy to find the right variables to instantiate and eliminate. It is also easy to find a variable order for which mini buckets produces good quality lower bounds. A natural question is whether it is possible to apply similar ideas to not so well structured problems. The answer is that it is often possible, although we need to rely on more naive and consequently less efficient exploitation of the problems' structure. In this Section we support our claim by reporting additional experimental results on different benchmarks. In particular, we consider *spot5* and DIMACS instances. *Spot5* instances are optimization problems taken from the scheduling of an earth observation satellite (Bensana, Lemaitre, & Verfaillie, 1999). The DIMACS benchmark contains SAT instances from several domain. Since we are concerned with optimization tasks, we have selected some unsatisfiable instances and solved the Max-SAT task (i.e, given an unsatisfiable SAT instance, find the maximum number of clauses that can be simultaneously satisfied), which can be modeled as a WCSP (de Givry, Larrosa, Meseguer, & Schiex, 2003). We consider *aim* instances (artificially generated random 3-SAT), *pret* (graph coloring), *ssa* and *bf* (circuit fault analysis).

Figure 8 shows the constraint graph of one instance of each domain, as visualized by LEDA graph editor. It can be observed that these graphs do not have an obvious pattern to be exploited. Thus, we have to use variable elimination techniques in a more naive way. We solve the problems with the generic WCSP solver TOOLBAR[7] (TB). It performs a depth-first branch-and-bound search and it is enhanced with general-purpose dynamic variable and value ordering heuristics. We modified TOOLBAR to combine search and variable elimination as follows: at an arbitrary subproblem, every variable with degree less than 3 is eliminated. Only when all the variables have degree larger than or equal to 3, an unassigned variable is heuristically selected and each of its domain values are heuristically ordered and sequentially instantiated. The process is recursively applied to each of the subproblems. Note that this is a generic version of the HYB algorithm where the decision of which variables are instantiated and which variables are eliminated is left to a heuristic, instead of establishing it by hand. We will refer to this implementation as $TB_{HYB}$. TOOLBAR offers a variety of lower bounds based on different forms of local consistency (Larrosa & Schiex, 2003). One of them, *directional arc consistency* (DAC*), is essentially equivalent to mini-buckets of size 2 and, therefore, similar in spirit to the lower bound computed by HYB. However, unlike HYB where mini-buckets are executed only once as a pre-process, TOOLBAR executes DAC* at every search state, subject to the current subproblem. It has been shown by Kask (2000) that this approach is generally more efficient. The other main difference with respect HYB, is that TOOLBAR executes DAC* subject to an arbitrary variable ordering (in HYB a good order was identified from the problem structure). Other lower bounds available in TOOLBAR are *node consistency* (NC*) which is weaker than DAC*, and *full directional arc consistency* (FDAC*) which can be seen as a (stronger) refinement of DAC*. We have experimented with four algorithms: $TB^{NC*}$, $TB^{DAC*}$, $TB_{HYB}^{DAC*}$ and $TB_{HYB}^{FDAC*}$, where $A^B$ denotes algorithm $A$ with lower bound $B$.

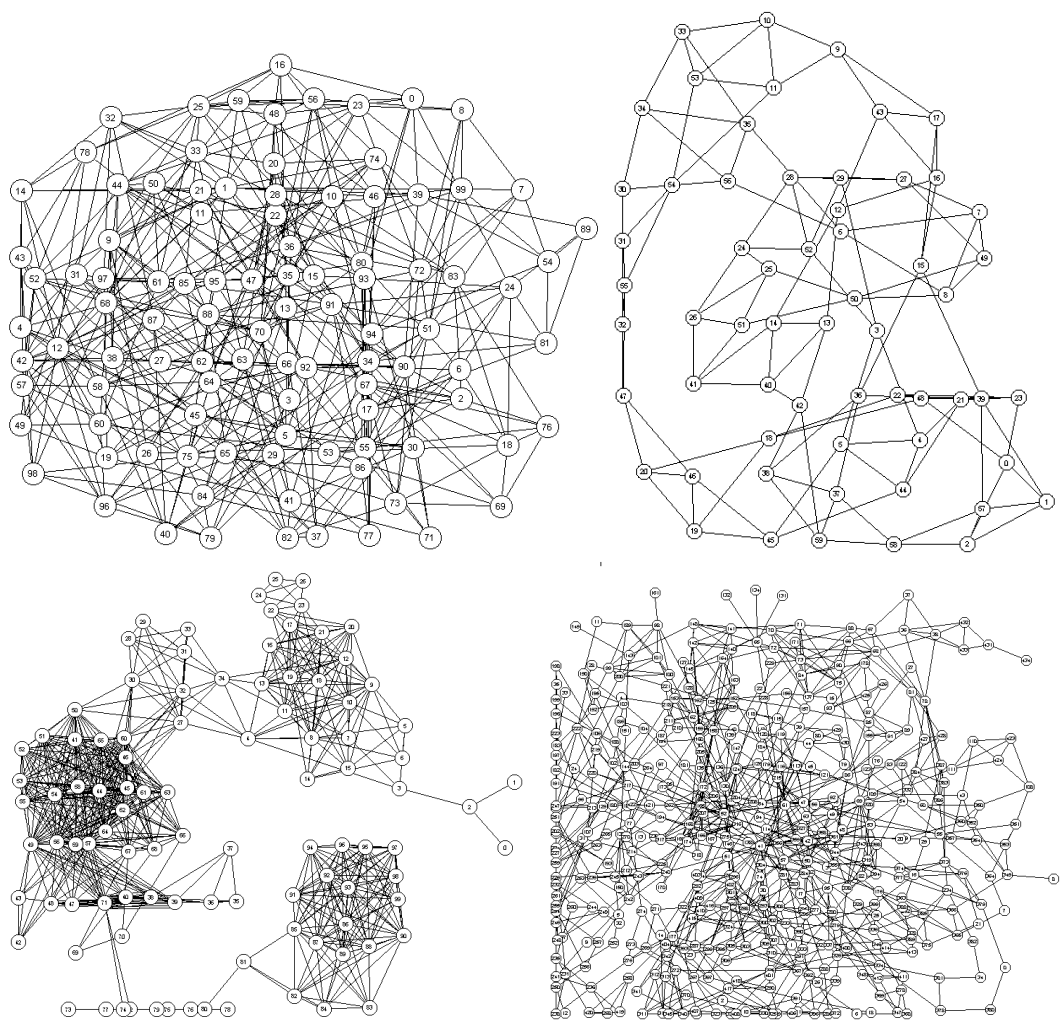

Figure 8: Constraint graph of four WCSP instances. From the top-left corner, clockwise, *aim-100-1-6-no-1*, *pret60-25*, *ssa0432-003* and *Spot5-404*.

---

7. Available at http://carlit.toulouse.inra.fr/cgi-bin/awki.cgi/SoftCSP.

Most *spot5* instances are too difficult for TOOLBAR. Therefore, we decreased their size by letting TOOLBAR make a sequence of $k$ greedy assignments driven by its default variable and value ordering heuristics. The result is a subproblem with $k$ less variables. In the following, $I_k$ denotes instance $I$ where $k$ variables have been greedily assigned by TOOLBAR with default parameters.

Table 9 reports the result of these experiments. The first column indicates the instances and subsequent columns indicate the CPU time (in seconds) required by the different algorithms. A time limit of 3600 seconds was set up for each execution. It can be observed that TOOLBAR with the weakest lower bound ($TB^{NC*}$) is usually the most inefficient alternative. It cannot solve any of the *spot5* instances and also fails with several *aim* and *ssa* instances. When TOOLBAR is enhanced with a mini buckets lower bound ($TB^{DAC*}$) all *spot5* problems are solved. In the other domains, the new lower bound does not produce a significant effect. When we further add variable elimination ($TB_{HYB}^{DAC*}$) all the problems are solved. In general, there is a clear speed-up. The worst improvements are in the *pret* instances where the time is divided by a factor of 2 and the best ones are obtained in the *spot5* $503_{40}$ and *ssa*7552-158 instances which are solved instantly. Typical speed-ups range from 5 to 10. Finally, we observe that the addition of the stronger lower bound ($TB_{HYB}^{FDAC*}$) has a limited effect in these problems. Only the execution of instance *ssa*7552-038 is clearly accelerated. Therefore, from these experiments we can conclude that the main techniques that we used to solve the still-life problem can also be successfully applied to other domains.

## 7. Conclusions

In this paper we have studied the applicability of variable elimination to the problem of finding *still-lifes*. Finding still-lifes is a challenging problem and developing new solving techniques is an interesting task *per se*. Thus, the first contribution of this paper is the observation that plain variable elimination (i.e, BE) is competitive in practice and provides time complexity exponentially better than search-based approaches. Besides, we have developed an algorithm with which we have been able to solve up to the $n = 20$ instance, with which we clearly improved previous results. The second contribution of the paper has a deeper insight. Our algorithm uses recent techniques based on variable elimination. Since these techniques are little known and rarely applied in the constraints community, the results presented in this paper add new evidence of their potential. We have also shown that variable elimination can be used beyond the academic still-life problem by providing experimental results in some unstructured realistic problems from different domains.

| Problem | $TB_{NC*}$ | $TB_{DAC*}$ | $TB_{HYB}^{DAC*}$ | $TB_{HYB}^{FDAC*}$ |
|---|---|---|---|---|
| Spot5 $404_0$ | - | 242 | 40 | 40 |
| Spot5 $408_{100}$ | - | 314 | 48 | 43 |
| Spot5 $412_{200}$ | - | 223 | 47 | 42 |
| Spot5 $414_{260}$ | - | 1533 | 221 | 139 |
| Spot5 $503_{40}$ | - | 546 | 0 | 0 |
| Spot5 $505_{120}$ | - | 3353 | 84 | 84 |
| Spot5 $507_{200}$ | - | 204 | 58 | 42 |
| Spot5 $509_{240}$ | - | 684 | 166 | 121 |
| aim-100-1-6-no-1 | - | - | 1665 | 1427 |
| aim-100-1-6-no-2 | - | - | 707 | 571 |
| aim-100-1-6-no-3 | - | - | 1960 | 1627 |
| aim-100-1-6-no-4 | - | - | 2716 | 2375 |
| aim-100-2-0-no-1 | 2516 | 2007 | 830 | 583 |
| aim-100-2-0-no-2 | 1191 | 931 | 479 | 285 |
| aim-100-2-0-no-3 | 1222 | 850 | 319 | 278 |
| aim-100-2-0-no-4 | 2162 | 1599 | 738 | 600 |
| bf0432-007 | - | - | 1206 | 1312 |
| pret60-25 | 110 | 120 | 49 | 56 |
| pret60-40 | 110 | 120 | 48 | 56 |
| ssa0432-003 | 22 | 22 | 5 | 5 |
| ssa2670-141 | - | - | 749 | 767 |
| ssa7552-038 | - | - | 20 | 2 |
| ssa7552-158 | - | - | 0 | 1 |

Figure 9: Experimental results in some WCSP instances with four different algorithms. Each column reports CPU time in seconds. Symbol - indicates that a time limit of 3600 seconds has been reached.

## Acknowledgments

The authors are grateful to Barbara Smith, Neil Yorke-Smith and the anonymous reviewers for their useful comments at different stages of the work reported in this article. Marti Sanchez kindly made the plots in Figure 8. This research has been funded by the Spanish CICYT under project TIC2002-04470-C03-01.